\documentclass[conference]{IEEEtran}
\IEEEoverridecommandlockouts
\usepackage{cite}
\usepackage{amsmath,amssymb,amsfonts}
\usepackage{algorithmic}
\usepackage{graphicx}
\usepackage{textcomp}
\usepackage{xcolor}
\usepackage[normalem]{ulem}
\def\BibTeX{{\rm B\kern-.05em{\sc i\kern-.025em b}\kern-.08em
    T\kern-.1667em\lower.7ex\hbox{E}\kern-.125emX}}

\newenvironment{ijcnn}{\color{black}}

\begin{document}

\title{Compensating Supervision Incompleteness with Prior Knowledge in Semantic Image Interpretation}

\author{\IEEEauthorblockN{Ivan Donadello}
\IEEEauthorblockA{\textit{Fondazione Bruno Kessler}\\
Trento, Italy \\
\texttt{donadello@fbk.eu}}
\and
\IEEEauthorblockN{Luciano Serafini} 
\IEEEauthorblockA{\textit{Fondazione Bruno Kessler}\\
Trento, Italy \\
\texttt{serafini@fbk.eu}}
}

\maketitle

\def\A{\mathcal{A}}
\def\abv{\mathsf{above}}
\def\animal{\mathsf{Animal}}
\def\ar{\mathit{ar}}
\def\arm{\mathsf{Arm}}
\def\arity{\alpha}
\def\below{\mathsf{below}}
\def\bicycle{\mathsf{Bicycle}}
\def\body{\mathsf{Body}}
\def\brown{\mathsf{Brown}}
\def\brownHorse{\mathsf{BrownHorse}}
\def\be{\mathbf{e}}
\def\bx{\mathbf{x}}
\def\by{\mathbf{y}}
\def\bt{\mathbf{t}}
\def\bu{\mathbf{u}}
\def\bus{\mathsf{Bus}}
\def\bv{\mathbf{v}}
\def\bw{\mathbf{w}}
\def\C{\mathcal{C}}
\def\camera{\mathsf{Camera}}
\def\car{\mathsf{Car}}
\def\carry{\mathsf{carry}}
\def\cat{\mathsf{Cat}}
\def\coach{\mathsf{Coach}}
\def\cone{\mathsf{Cone}}
\def\D{\mathcal{D}}
\def\dog{\mathsf{Dog}}
\def\drive{\mathsf{drive}}
\def\dress{\mathsf{Dress}}
\def\E{\mathcal{E}}
\def\eat{\mathsf{eat}}
\def\elephant{\mathsf{Elephant}}
\def\face{\mathsf{Face}}
\def\female{\mathsf{Female}}
\def\Fn{\mathcal{F}}
\def\furia{\mathsf{furia}}
\def\G{\mathcal{G}}
\def\GG{\mathbb{G}}
\def\grass{\mathsf{Grass}}
\def\GT{\langle\K,\pG\rangle}
\def\gtsii{\mathcal{T}_{\mathrm{SII}}}
\def\gtw{\mathcal{T}_{\mathrm{prior}}}
\def\gtwo{\mathcal{T}_{\mathrm{expl}}}
\def\handlebar{\mathsf{Handlebar}}
\def\has{\mathsf{has}}
\def\haspart{\mathsf{hasPart}}
\def\head{\mathsf{Head}}
\def\headlight{\mathsf{Headlight}}
\def\headOf{\mathsf{headOf}}
\def\hit{\mathsf{hit}}
\def\hold{\mathsf{hold}}
\def\horse{\mathsf{Horse}}
\def\human{\mathsf{Human}}
\def\I{\mathcal{I}}
\def\imp{\rightarrow}
\def\ir{\mathit{ir}}
\def\isa{\sqsubseteq}
\def\jacket{\mathsf{Jacket}}
\def\john{\mathsf{john}}
\def\johnny{\mathsf{johnny}}
\def\julia{\mathsf{julia}}
\def\K{\mathcal{K}}
\def\kick{\mathsf{kick}}
\def\KB{\mathcal{KB}}
\def\ksii{\mathcal{K}_{\mathrm{SII}}}
\def\kw{\mathcal{K}_{\mathrm{prior}}}
\def\kwo{\mathcal{K}_{\mathrm{expl}}}
\def\Lng{\mathcal{PL}} 
\def\leg{\mathsf{Leg}}
\def\legOf{\mathsf{legOf}}
\def\legs{\mathsf{Legs}}
\def\leftOf{\mathsf{left\ of}}
\def\limb{\mathsf{Limb}}
\def\Loss{\mathscr{L}}
\def\male{\mathsf{Male}}
\def\mammal{\mathsf{Mammal}}
\def\motorcycle{\mathsf{Motorcycle}}
\def\muzzle{\mathsf{Muzzle}}
\def\Nat{\mathbb{N}}
\def\near{\mathsf{near}}
\def\nextTo{\mathsf{next\ to}}
\def\O{\mathcal{O}}
\def\on{\mathsf{on}}
\def\ovr{\mathsf{over}}
\def\P{\mathcal{P}}
\def\partof{\mathsf{partOf}}
\def\pascalpart{\textsc{PASCAL-Part}}
\def\parkOn{\mathsf{park\ on}}
\def\person{\mathsf{Person}}
\def\pgsii{\hat{\mathcal{G}}_{\mathrm{SII}}}
\def\pG{\hat{\mathcal{G}}}
\def\phyObj{\mathsf{PhyObj}}
\def\pics{\mathit{Pics}}
\def\pizza{\mathsf{Pizza}}
\def\PL{\mathcal{PL}}
\def\poftype{\mathsf{ptype}}
\def\pOmega{\hat{\Omega}}
\def\Pr{\mathcal{P}}
\def\precision{\mathit{prec}}
\def\R{\mathbb{R}}
\def\RBox{\mathcal{R}}
\def\recall{\mathit{rec}}
\def\ride{\mathsf{ride}}
\def\rightOf{\mathsf{right\ of}}
\def\SROIQ{\mathcal{SROIQ}}
\def\shirt{\mathsf{Shirt}}
\def\saddle{\mathsf{Saddle}}
\def\sibling{\mathsf{sibl}}
\def\sitBehind{\mathsf{sit\ behind}}
\def\sleepOn{\mathsf{sleep\ on}}
\def\standOn{\mathsf{stand\ on}}
\def\street{\mathsf{Street}}
\def\T{\mathcal{T}}
\def\tab{\mathsf{Table}}
\def\tail{\mathsf{Tail}}
\def\tailof{\mathsf{tailOf}}
\def\tallerThan{\mathsf{taller\ than}}
\def\tensorflow{\textsc{TensorFlow}$^{TM}$}
\def\term{\mathit{term}}
\def\train{\mathsf{Train}}
\def\truck{\mathsf{Truck}}
\def\wear{\mathsf{wear}}
\def\with{\mathsf{with}}
\def\wheel{\mathsf{Wheel}}
\def\under{\mathsf{under}}
\def\vegtable{\mathsf{Vegetable}}
\def\vehicle{\mathsf{Vehicle}}

\newcommand{\argmin}{\mathop{\mathrm{argmin}}}
\newcommand{\argmax}{\mathop{\mathrm{argmax}}}

\newtheorem{definition}{Definition}
\newenvironment{dedication}
     {\begin{quotation}\begin{center}\begin{em}}
     {\par\end{em}\end{center}\end{quotation}}
\begin{abstract}
Semantic Image Interpretation is the task of extracting a structured semantic description from images. This requires the detection of \emph{visual relationships}: triples \emph{$\langle$subject,~relation,~object$\rangle$} describing a semantic relation between a subject and an object. A pure supervised approach to visual relationship detection requires a complete and balanced training set for all the possible combinations of \emph{$\langle$subject,~relation,~object$\rangle$}. However, such training sets are not available and would require a prohibitive human effort. This implies the ability of predicting triples which do not appear in the training set. This problem is called \emph{zero-shot learning}. State-of-the-art approaches to zero-shot learning exploit similarities among relationships in the training set or external linguistic knowledge. In this paper, we perform zero-shot learning by using Logic Tensor Networks, a novel Statistical Relational Learning framework that exploits both the similarities with other seen relationships and background knowledge, expressed with logical constraints between subjects, relations and objects. The experiments on the Visual Relationship Dataset show that the use of logical constraints outperforms the current methods. This implies that background knowledge can be used to alleviate the incompleteness of training sets. 
\end{abstract}
\section{Introduction}\label{sec:intro}
Semantic Image Interpretation (SII) \cite{neumann2008onscene} concerns the automatic extraction of high-level information about the content of a visual scene. This information regards the objects in the scene, their attributes and the relations among them. Formally, SII extracts the so-called \emph{scene graph} \cite{krishna2016visualgenome} from a picture: the labelled nodes refer to objects in the scene and their attributes, the labelled edges regard the relations between the corresponding nodes. SII enables important applications on an image content that a coarser image analysis (e.g., the object detection) does not allow. For example, the \emph{visual question answering} (answering to natural language questions about the image content), the \emph{image captioning} (generating a natural language sentence describing the image content), the \emph{complex image querying} (retrieving images using structured queries about the image content) or the \emph{robot interaction} (different configurations of objects allow different actions for a robot). The visual relationship detection (VRD) is a special instance of scene graph construction. Indeed, a visual relationship is a triple $\langle subject, predicate, object \rangle$ where the subject and the object are the labels (or semantic classes) of two bounding boxes in the image. The predicate is the label regarding the relationship between the two bounding boxes. The construction of a scene graph from visual relationships is pretty forward. The subject and the object are nodes in the graph with the corresponding labels, the predicate is a labelled edge from the subject node to the object node.

Visual relationships are mainly detected with supervised learning techniques \cite{yu17vrd}. These require large training sets of images annotated with bounding boxes and relationships \cite{lu2016visual,krishna2016visualgenome}. However, a complete and detailed annotation is not possible due to the high human effort of the annotators. For example, a person riding a horse can be annotated with the relations $\on$, $\ride$ (person subject) or $\below$, $\carry$ (horse subject). Thus, many types of relationships are not in the training set but can appear in the test set. The task of predicting visual relationships with never seen instances in the training phase is called \emph{zero-shot learning} \cite{Lampert2014Attribute}. This can be achieved by exploiting the similarity with the triples in the training set or using a high-level description of the relationship. For example, the fact that people ride elephants can be derived from a certain similarity between elephants and horses and the (known) fact that people can ride horses. This is closer to human learning with respect to supervised learning. Indeed, humans are able to both generalize from seen or similar examples and to use their background knowledge to identifying never seen relationships \cite{Lampert2014Attribute}. This background knowledge can be linguistic information about the subject/object and predicate occurrences \cite{lu2016visual,Zhang2017VtransE,yu17vrd} or logical constraints \cite{serafini2017learning,donadello2017ijcai}. Logical knowledge is very expressive and it allows us to explicitly state relations between subjects/objects and predicates. For example, the formula $\forall x,y( ride(x,y) \rightarrow elephant(y) \vee horse(y))$ states that the objects of the riding relations are elephants and horses. Other formulas can state that horses and elephants cannot ride and this avoids wrong predictions.

In this paper, we address the zero-shot learning problem by using Logic Tensor Networks (LTNs) \cite{Serafini2016ltn} for the detection of unseen visual relationships. LTNs is a Statistical Relational Learning framework that learns from relational data (exploiting the similarities with already seen triples) in presence of logical constraints. The results on the Visual Relationship Dataset show that the joint use of logical knowledge and data outperforms the state-of-the-art approaches based on data and/or linguistic knowledge. These promising results show that logical knowledge is able to counterbalance the incompleteness of the datasets due to the high annotation effort: this is a significant contribution to the zero-shot learning. LTNs have already been exploited for SII \cite{serafini2017learning,donadello2017ijcai}. However, these works are preliminary as they focus only on the part-whole relation (\pascalpart\ dataset \cite{chen14PascalPart}). We extend these works with the following contributions:
\begin{itemize}
\item We conduct the experiments on the the Visual Relationship Dataset (VRD) \cite{lu2016visual}, a more challenging dataset that contains 70 binary predicates.
\item We introduce new additional features for pairs of bounding-boxes that capture the geometric relations between bounding boxes. These features are necessary as they drastically improve the performance.
\item We perform a theoretical analysis of the drawbacks of using a loss function based on t-norms. Therefore, we introduce a new loss function based on the harmonic mean among the truth values of the axioms in the background knowledge. 
\end{itemize}
The effectiveness of the new features and the new mean-based loss function is proved with some ablation studies.
\section{Related Work}
\label{sec:related-work}
The detection of visual relationships in images is tightly connected to Semantic Image Interpretation (SII) \cite{neumann2008onscene}. SII extracts a graph \cite{krishna2016visualgenome} that describes an image semantic content: nodes are objects and attributes, edges are relations between nodes. In \cite{neumann2008onscene} the SII graph (i.e., the visual relationships) is generated with deductive reasoning and the low-level image features are encoded in a knowledge base with some logical axioms. In \cite{Peraldi2009formalizing} abductive reasoning is used. However, writing axioms that map features into concepts/relations or defining the rules of abduction requires a high engineering effort, and dealing with the noise of the object detectors could be problematic. Fuzzy logic \cite{hajek2001metamathematics} deals with this noise. In \cite{Hudelot2008Fuzzy,Hudelot2014FCA} a fuzzy logic ontology of spatial relations and an algorithm (based on morphological and logical reasoning) for building SII graphs are proposed. These works are limited to spatial relations. In \cite{xu2017sceneGraph} the SII graph is built with an iterative message passing algorithm where the information about the objects maximizes the likelihood of the relationships and vice versa. \begin{ijcnn}
In \cite{Zellers18motifs} a combination of Long Short-Term Memories (LSTMs) is exploited. The first LSTM encodes the context given by the detected bounding boxes. This context is used for classifying both objects and relations with two different LSTMs without external knowledge. Here, only one edge is allowed between two nodes (as in \cite{yu17vrd,xu2017sceneGraph}). This assumption of mutual exclusivity between predicates does not hold in real life, e.g., if a person rides a horse then is on the horse. LTNs instead allow multiple edges between nodes. In \cite{donadello2014mixing} a clustering algorithm integrates low-level and semantic features to group parts belonging to the same whole object or event. Logical reasoning is applied for consistency checking. However, the method is tailored only on the part-whole relation.\end{ijcnn}

Other methods start from a fully connected graph whose nodes and edges need to be labelled or discarded according to an energy minimization function. In \cite{kulkarni2011baby} the graph is encoded with a Conditional Random Field (CRF) and potentials are defined by combining the object detection score with geometric relations between objects and text priors on the types of objects. Also in \cite{chen14PascalPart} the scene graph is encoded with a CRF and the work leverages the part-whole relation to improve the object detection. These works do not consider logical knowledge. In \cite{chen2012understanding} the energy function combines visual information of the objects with logical constraints. However, this integration is hand-crafted and thus difficult to extend to other types of constraints.

A \emph{visual phrase} \cite{sadeghi2011recognition} is the prototype of a visual relationship. Here a single bounding box contains both subject and object. However, training an object detector for every possible triple affects the scalability. The \emph{visual semantic role labelling} \cite{gupta2015visual} is a generalization of detecting visual relationships. This task generates a set of tuples, such as: $\langle predicate, \{\langle role_1, label_1 \rangle, \ldots, \langle role_N, label_N\rangle\}\rangle$, where the roles are entities involved in the predicate, such as, subject, object, tool or place. However, the work is preliminary and limits the role of subject only to people.

Other works exploit deep learning. In \cite{Dai2017Detecting} the visual relationships are detected with a Deep Relational Network that exploits the statistical dependencies between relationships and the subjects/objects. In \cite{Liang2017Deep} a deep reinforcement learning framework to detect relationships and attributes is used. In \cite{Li2017ViP} a message passing algorithm is developed to share information about the subject, object and predicate among neural networks. In \cite{Zhang2017VtransE} the visual relationship is the translating vector of the subject towards the object in an embedding space. In \cite{yin2018zoom} an end-to-end system exploits the interaction of visual and geometric features of the subject, object and predicate. The end-to-end system in \cite{Zhang2017PPR-FCN} exploits weakly supervised learning (i.e., the supervision is at image level). LTNs exploit the combination of the visual/geometric features of the subject/object with additional background knowledge.

Background knowledge is also exploited in a joint embedding with visual knowledge. In \cite{ramanathan2015learning} the exploited logical constraints are implication, mutual exclusivity and type-of. In \cite{lu2016visual} the background knowledge is a word embedding of the subject/object labels. The visual knowledge consists in the features of the union of the subject and object bounding boxes. In \cite{tresp2017improvingVRD} the background knowledge is statistical information (learnt with statistical link prediction methods \cite{nickel2016Review}) about the training set triples. Contextual information between objects is used also in \cite{Zhuang2017context,Peyre2017Weakly} with different learning methods. In \cite{yu17vrd} the background knowledge (from the training set and Wikipedia) is a probability distribution of a relationship given the subject/object. This knowledge drives the learning of visual relationships. These works do not exploit any type of logical constraints as LTNs do.
\section{Logic Tensor Networks}
\label{sec:ltn}
In the following we describe the basic notions of Logic Tensor Networks (LTNs) \cite{Serafini2016ltn}, whereas in Section \ref{sec:ltn_vrd} we present the novel contributions of our work evaluated in Section \ref{sec:evaluation}. LTNs are a statistical relational learning framework that combine Neural Networks with logical constraints. LTNs adopt the syntax of a First-Order predicate language $\Lng$ whose signature is composed of two disjoint sets $\C$ and $\Pr$ denoting constants and predicate symbols, respectively. We do not present LTNs function symbols as they are not strictly necessary for the detection of visual relationships. $\Lng$ allows LTNs to express visual relationships and a priori knowledge about a domain. E.g., the visual relationship $\langle person, ride, horse\rangle$ is expressed with the atomic formulas $\person(p_1)$, $\horse(h_1)$ and $\ride(p_1, h_1)$. Common knowledge is expressed through logical constraints, e.g., $\forall x,y (\ride(x,y) \rightarrow \neg \dog(x))$ states that dogs do not ride.

LTNs semantics deviates from the \textit{abstract} semantics of Predicate Fuzzy Logic \cite{hajek2001metamathematics} towards a \textit{concrete} semantics. Indeed, the interpretation domain is a subset of $\mathbb{R}^n$, i.e., constant symbols and closed terms are associated with a $n$-dimensional real vector. This vector encodes $n$ numerical features of an object, such as, the confidence score of an object detector, the bounding box coordinates, local features, etc. Predicate symbols are interpreted as functions on real vectors to $[0,1]$. The interpretation of a formula is the degree of truth for that formula: higher values mean higher degrees of truth. LTNs use the term \emph{grounding} as synonym of logical interpretation in a ``real'' world. A grounding has to capture the latent correlation between the features of objects and their categorical/relational properties. Let $\arity(s)$ denote the arity of a predicate symbol $s$.
\begin{definition} 
An $n$-\emph{grounding} $\G$ (with $n \in\Nat$, $n > 0$), or simply \emph{grounding}, for a First-Order Language $\Lng$ is a function from the signature of $\Lng$ that satisfies the conditions:
\begin{enumerate}
\item $\G(c)\in\R^n$, for every $c\in\C$;
\item $\G(P)\in \R^{n\cdot \arity(P)}\longrightarrow [0,1]$, for every $P\in\Pr$. 
\end{enumerate}
\end{definition}
\noindent Given a grounding $\G$ and let $\term(\Lng) = \{t_1, t_2, t_3, \ldots\}$ be the set of closed terms of $\Lng$, the semantics of atomic formulas is inductively defined as follows:
\begin{align}\label{eq:grPf}
\G(P(t_1,\dots,t_m)) 		& = \G(P)(\G(t_1),\dots,\G(t_m)).
\end{align}
\noindent The semantics for non-atomic formulas is defined according to t-norms functions used in Fuzzy Logic. If we take, for example, the \L{}ukasiewicz t-norm, we have:
\begin{align}
\label{eq:grLuk}
\G(\phi\imp\psi) 		& = \min(1,1-\G(\phi)+\G(\psi))			\nonumber\\
\G(\phi\wedge\psi) 		& = \max(0,\G(\phi)+\G(\psi) - 1) 		\nonumber\\ 
\G(\phi\vee\psi) 		& = \min(1,\G(\phi)+\G(\psi)) 			\nonumber\\ 
\G(\neg\phi) 			& = 1-\G(\phi).
\end{align}
\noindent
The semantics for quantifiers differs from the semantics of standard Fuzzy Logic. Indeed, the interpretation of $\forall$ leads the definition:
\begin{equation}
\G(\forall x \phi(x)) = \inf\{\G(\phi(t)) | t \in \term(\Lng)\}.
\end{equation}
This definition does not tolerate exceptions. E.g., the presence of a dog riding a horse in a circus would falsify the formula that dogs do not ride. LTNs handle these outliers giving a higher truth-value to the formula $\forall x \phi(x)$ if many examples satisfy $\phi(x)$. This is in the spirit of SII as in a picture (due to occlusions or unexpected situations) some common logical constraints are not always respected.
\begin{definition}
\label{def:grUniv}
Let $mean_p(x_1,\dots,x_d) = \left(\frac{1}{d}\sum_{i=1}^d x_i^{p}\right)^{\frac{1}{p}}$, with $p\footnote{The popular mean operators (arithmetic, geometric and harmonic mean) are obtained by setting p = 1, 2, and -1, respectively.} \in \mathbb{Z}$, $d \in \mathbb{N}$, the grounding for $\forall x \phi(x)$ is
\begin{multline}
\label{eq:grUniv}
\G(\forall x \phi(x)) = \\
\lim_{d\rightarrow \vert \term(\Lng) \vert }mean_p(\G(\phi(t_1)), \ldots ,\G(\phi(t_d))).
\end{multline}
\end{definition}
\noindent The grounding of a quantified formula $\forall x \phi(x)$ is the mean of the $d$ groundings of the quantifier-free formula $\phi(x)$.

A suitable function for a grounding should preserve some form of $\textit{regularity}$. Let $b \in \C$ refer to a bounding box constant containing a horse. Let $\bv = \G(b)$ be its feature vector, then it holds that $\G(\horse)(\bv)\approx 1$. Moreover, for every bounding box with feature vector $\bv'$ similar to $\bv$, $\G(\horse)(\bv')\approx 1$ holds. These functions are learnt from data\footnote{Some groundings can be manually defined with some rules \cite{bloch2005fuzzySpatial}. However, for some predicates this could be time consuming or inaccurate \cite{donadello2017ijcai}.} by tweaking their inner parameters in a training process. 
The grounding for \textbf{predicate symbols} is a generalization of a neural tensor network: an effective architecture for relational learning \cite{nickel2016Review}. Let $b_1,\ldots,b_m \in \C$ with feature vectors $\bv_i = \G(b_i) \in \mathbb{R}^n$, with $i = 1 \ldots m$, and $\bv=\left<\bv_1;\dots;\bv_m\right>$ is a $mn$-ary vector given by the vertical stacking of each vector $\bv_i$. The grounding $\G(P)$ of an $m$-ary predicate $P(b_1,\ldots,b_m)$ is:
\begin{align}
\label{eq:grP}
\G(P)(\bv) = \sigma\left(u^\intercal_P\tanh\left(\bv^\intercal W_P^{[1:k]}\bv + V_P\bv + b_P\right)\right)
\end{align}
with $\sigma$ the sigmoid function. The parameters for $P$ are: $u_P \in \R^{k}$, a 3-D tensor $W_P^{[1:k]} \in \R^{k \times mn\times mn}$, $V_P \in \R^{k\times mn}$ and $b_P \in \R^{k}$. The parameter $u_P$ computes a linear combination of the quadratic features returned by the tensor product. With Equations \eqref{eq:grPf} and \eqref{eq:grP} the grounding of a complex LTNs formula can be computed by first computing the groundings of the closed terms and the atomic formulas contained in the complex formula. Then, these groundings are combined using a specific t-norm, see Equation \eqref{eq:grLuk}.

Learning the groundings involves the optimization of the truth values of the formulas in a LTNs knowledge base, a.k.a. \emph{grounded theory}. A \emph{partial grounding} $\pG$ is a grounding defined on a subset of the signature of $\Lng$. A grounding $\G$ for $\Lng$ is a \emph{completion} of $\pG$ (in symbols $\pG \subseteq \G$) if $\G$ coincides with $\pG$ on the symbols where $\pG$ is defined.

\begin{definition}
A \emph{grounded theory GT} is a pair $\GT$ with $\K$ a set of closed formulas and $\pG$ a partial grounding.
\end{definition}
\begin{definition}
A grounding $\G$ satisfies a grounded theory $\GT$ if $\pG \subseteq \G$ and $\G(\phi)=1$, for all $\phi\in\K$. A grounded theory $\GT$ is \emph{satisfiable} if there exists a grounding $\G$ that satisfies $\GT$. 
\end{definition}
\noindent According to the above definition, the satisfiability of $\GT$ can be obtained by searching for a grounding $\G$ that extends $\pG$ such that \emph{every} formula in $\K$ has value 1. When a grounded theory is not satisfiable a user can be interested in a degree of satisfaction of the GT.

\begin{definition}
Let $\GT$ be a grounded theory, the \emph{best satisfiability problem} amounts at searching an extension $\G^*$ of $\pG$ in $\GG$ (the set of all possible groundings) that maximizes the truth value of the conjunction of the formulas in $\K$:
\begin{equation}
\label{eq:besta}
\G^* = \argmax_{\pG \subseteq \G \in \GG}\G\left(\bigwedge_{\phi \in \K}\phi\right).
\end{equation}
\end{definition}

\noindent The best satisfiability problem is an optimization problem on the set of parameters to be learned. Let $\Theta = \{W_P,V_P,b_P,u_P \mid P\in\Pr\}$ be the set of parameters. Let $\G(\cdot|\Theta)$ be the grounding obtained by setting the parameters of the grounding functions to $\Theta$. The best satisfiability problem tries to find the best set of parameters $\Theta$:
\begin{equation}
\label{eq:ltn_loss}
\mbox{$\Theta^*=\argmax_{\Theta}\G\left(\left.\bigwedge_{\phi \in \K}\phi\right|\Theta\right) - \lambda\lVert\Theta\rVert_2^2$}
\end{equation}
with $\lambda\lVert\Theta\rVert_2^2$ a regularization term.
\section{LTNs for Visual Relationship Detection}\label{sec:ltn_vrd}
\begin{ijcnn}Similarly to \cite{donadello2017ijcai}, we encode the problem of detecting visual relationship with LTNs. However, the problem here is more challenging as the VRD contains many binary predicates and not only the $\partof$ as in the \pascalpart\ dataset. In the following we describe the novel contributions of our work.\end{ijcnn}

Let $\pics$ be a dataset of images. Given a picture $p \in \pics$, let $B(p)$ the corresponding set of bounding boxes. Each bounding box in $B(p)$ is annotated with a set of labels that describe the contained physical object. Pairs of bounding boxes are annotated with the semantic relations between the contained physical objects. Let $\Sigma_{\text{SII}}=\left<\P,\C\right>$ be a $\Lng$ signature where $\Pr=\Pr_1 \cup \Pr_2$ is the set of predicates. $\Pr_1$ is the set of unary predicates that are the \emph{object types} (or \emph{semantic classes}) used to label the bounding boxes, e.g., $\Pr_1 = \{\horse, \person, \shirt,\pizza, \dots\}$. The set $\Pr_2$ contains binary predicates used to label pairs of bounding boxes, e.g., $\Pr_2 = \{\ride,\on,\wear, \eat, \dots\}$. Let $\C=\bigcup_{p\in\pics}B(p)$ be the set of constants for all the bounding boxes in the dataset $\pics$. However, the information in $\pics$ is incomplete: many bounding boxes (or pairs of) have no annotations or even some pictures have no annotations at all. Therefore, LTNs is used to exploit the information in $\pics$ to complete the missing information, i.e., to predict the visual relationships. As in \cite{donadello2017ijcai}, we encode $\pics$ with a grounded theory $\gtsii =\langle\ksii,\pgsii\rangle$ described in the following. The LTNs knowledge base $\ksii$ encodes the bounding box annotations in the dataset and some background knowledge about the domain. The task is to complete the partial knowledge in $\pics$ by finding a grounding $\G^*_\mathrm{SII}$, that extends $\pgsii$, such that:
$$
\begin{array}{lcl}
\G^*_\mathrm{SII}(C(b)) 			& \mapsto &	[0, 1]\\
\G^*_\mathrm{SII}(R(b_1, b_2)) 	& \mapsto &	[0, 1]
\end{array}
$$
for every unary ($C$) and binary ($R$) predicate symbol and for every (pair of) bounding box in the dataset.

\subsection{The Knowledge Base $\ksii$}
\label{sec:kb_sii}
The knowledge base $\ksii$ contains positive and negative examples (used for learning the grounding of the predicates in $\Pr$) and the background knowledge. The \textbf{positive examples} (taken from the annotations in $\pics$) for a semantic class $C$ are the atomic formulas $C(b)$, for every bounding box $b$ labelled with class $C \in \Pr_1$ in $\pics$. The positive examples for a relation $R$ are the atomic formulas $R(b_1, b_2)$, for every pair of bounding boxes $\left<b_1, b_2\right>$ labelled with the binary relation $R \in \Pr_2$ in $\pics$. \begin{ijcnn}Regarding the \textbf{negative examples}, for a semantic class $C$ we consider the atomic formulas $\neg C(b)$, for every bounding box $b$ not labelled with $C$. The negative examples for a relation $R$ are the atomic formulas $\neg R(b_1, b_2)$, for every pair of bounding boxes $\left<b_1, b_2\right>$ not labelled with $R$.\end{ijcnn}


Regarding the \textbf{background knowledge}, we manually build the logical constraints. We focus on the \emph{negative domain and range constraints} that list which are the semantic classes that cannot be the subject/object for a predicate. E.g., clothes cannot drive. For every unary predicate $\dress$ in $\P_1$ that refers to a dress, the constraint $\forall xy(\drive(x,y) \imp \neg \dress(x))$ is added to $\mathcal{BK}$. In a large scale setting, these constraints can be retrieved by on-line linguistic resources such as FrameNet \cite{Baker1998framenet} and VerbNet \cite{Schuler2005verbnet} that provide the range and domain of binary relations through the so-called \emph{frames} data structure. Then, by applying mutual exclusivity between classes we obtain the negative domain and range constraints. This class of constraints brings to good performance. Experiments with other classes of constraints (such as IsA, mutual exclusivity, symmetry or reflexivity properties) are left as future work.

\subsection{The Grounding $\pgsii$}
\label{sec:gr_sii}
The grounding of each bounding box \textbf{constant} $b\in\C$ is a feature vector $\pgsii(b) = \bv_b \in \R^{|\Pr_1|+4}$ of \emph{semantic} and \emph{geometric} features:
\begin{multline}
\label{eq:grC}
\bv_b = \langle score(C_1,b),\dots,score(C_{|\Pr_1|},b)\\
x_0(b),y_0(b),x_1(b),y_1(b) \rangle,
\end{multline}
with $x_0(b),y_0(b),x_1(b),y_1(b)$ the coordinates of the top-left and bottom-right corners of $b$ and $score(C_i, b)$ is the classification score of an object detector for $b$ according to the class $C_i \in \Pr_1$. However, here we adopt the \emph{one-hot encoding}: the semantic features take value 1 in the position of the class with the highest detection score, 0 otherwise. The geometric features remain unchanged. The grounding for a pair of bounding boxes $\langle b_1, b_2 \rangle$ is the concatenation of the groundings of the single bounding boxes $\langle \bv_{b_1}: \bv_{b_2} \rangle$. However, when dealing with $n$-tuples of objects, adding new extra feature regarding geometrical joint properties of these $n$ bounding boxes improves the performance of the LTNs model. Differently from \cite{donadello2017ijcai}, we add more \emph{joint features}\footnote{All the features are normalized in the interval $[-1, 1]$.} that better capture the geometric interactions between bounding boxes:
\begin{multline}
\label{eq:grCC}
\bv_{b_1, b_2} = \langle \bv_{b_1}: \bv_{b_2}, ir(b_1, b_2), ir(b_2, b_1), \frac{area(b_1)}{area(b_2)}, \\ \frac{area(b_2)}{area(b_1)}, euclid\_dist(b_1, b_2), sin(b_1, b_2), cos(b_1, b_2) \rangle
\end{multline}
where:
\begin{itemize}
\item $\ir(b_1, b_2) = intersec(b_1, b_2)/area(b_1)$ is the inclusion ratio, see \cite{donadello2017ijcai};
\item $area(b)$ is the area of $b$;
\item $intersec(b_1, b_2)$ is the area of the intersection of $b_1, b_2$;
\item $euclid\_dist(b_1, b_2)$ is the Euclidean distance between the centroids of bounding boxes $b_1, b_2$;
\item $sin(b_1, b_2)$ and $cos(b_1, b_2)$ are the sine and cosine of the angle between the centroids of $b_1$ and $b_2$ computed in a counter-clockwise manner.
\end{itemize}
Regarding the \textbf{unary predicates} in $\P_1$, we adopt a rule-based grounding. Given a bounding box constant $b$, its feature vector $\bv_b=\left<v_1,\dots,v_{|\Pr_1|+4}\right>$, and a predicate symbol $C_i\in\Pr_1$, the grounding for $C_i$ is:
\begin{align}
\label{eq:grUnHot}
\pgsii(C_i)(\bv_b)=\left\{
\begin{array}{ll}
1 & \mbox{if } i = \argmax_{1 \leq l \leq |\Pr_1|}\bv_b^l \\ 
0 & \mbox{otherwise}.
\end{array}
\right. 
\end{align}
Regarding the \textbf{binary predicates} in $\P_2$, a rule-based grounding would require a different analysis for each predicate and could be inaccurate. Therefore, this grounding is learned from data by maximizing the truth values of the formulas in $\ksii$. The grounding of the \textbf{logical constraints} in $\ksii$ is computed by (i) instantiating a tractable sample of the constraints with bounding box constants belonging to the same picture. (ii) Computing the groundings of the atomic formulas of every instantiated constraint; (iii) combining the groundings of the atomic formulas according to the LTNs semantics; (iv) aggregating the groundings of every instantiated constraint according to the LTNs semantics of $\forall$.

\subsection{The Optimization of $\gtsii$}
\label{sec:opt_sii}
Equation \eqref{eq:ltn_loss} defines how to learn the LTNs parameters by maximizing the grounding of the conjunctions of the formulas in $\ksii$. Here we analyze some problems that can arise when the optimization is performed with the main t-norms:
\begin{LaTeXdescription}
\item[\L{}ukasiewicz t-norm] The satisfiability of $\ksii$ is given by: $\pgsii(\bigwedge_{\phi \in \ksii}\phi) = \max\{0, \sum_{\phi \in \ksii}\pgsii(\phi) - |\ksii| + 1\}$. Thus, the higher the number of formulas the higher their grounding should be to have a satisfiability value bigger than zero. However, even a small number of formulas in $\ksii$ with a low grounding value can lead the knowledge base satisfiability to zero.
\item[G\"odel t-norm] The satisfiability of $\ksii$ is the minimum of the groundings of all its formulas: $\pgsii(\bigwedge_{\phi \in \ksii}\phi) = \min\{\pgsii(\phi) | \phi \in \ksii \}$. Here the optimization process could get stuck in a local optimum. Indeed, a single predicate could be too difficult to learn, the optimizer tries to increase this value without any improvement and thus leaving out the other predicates from the optimization.
\item[Product t-norm] The satisfiability of $\ksii$ is the product of the groundings of all its formulas: $\pgsii(\bigwedge_{\phi \in \ksii}\phi) = \prod_{\phi \in \ksii}\pgsii(\phi)$. As $\ksii$ can have many formulas, the product of hundreds of groundings can result in a very small number and thus incurring in underflow problems.
\end{LaTeXdescription}
Differently from \cite{donadello2017ijcai}, we provide another definition of satisfiability. We use a mean operator in Equation \eqref{eq:ltn_loss} that returns a \emph{global satisfiability} of $\ksii$ avoiding the mentioned issues:
\begin{equation}
\label{eq:ltn_loss_sii}
\mbox{$\Theta^*=\argmax_{\Theta}mean_p\left(\pgsii(\phi|\Theta) | \phi \in \ksii \right) - \lambda\lVert\Theta\rVert_2^2$},
\end{equation}
with $p \in \mathbb{Z}$. Here, we avoid $p \geq 1$ as the obtained means are more influenced by the higher grounding values, that is, by the predicates easy to learn. These means return a too optimistic value of the satisfiability and this wrongly avoids the need of optimization. The computation of Equation \eqref{eq:ltn_loss_sii} is linear with respect to the number of formulas in $\ksii$.

\subsection{Post Processing}
Given a trained grounded theory $\gtsii$, we compute the set of groundings $\{\pgsii(\mathsf{r}(b,b'))\}_{\mathsf{r} \in \P_2}$, with $\langle b, b' \rangle$ a new pair of bounding boxes. Then, every grounding $\pgsii(\mathsf{r}(b,b'))$ is multiplied with a prior: the frequency of the predicate $\mathsf{r}$ in the training set. In addition, we exploit equivalences between the binary predicates (e.g., beside is equivalent to next to) to normalize the groundings. In the specific, $\pgsii(\mathsf{r_1}(b,b')) = \pgsii(\mathsf{r_2}(b,b')) = \max\{\pgsii(\mathsf{r_1}(b,b')), \pgsii(\mathsf{r_2}(b,b'))\}$, if $\mathsf{r_1}$ and $\mathsf{r_2}$ are equivalent.
\section{Experiments}\label{sec:evaluation}
\label{sec:results}
We conduct the experiments\footnote{The source code and the models are available at \mbox{https://github.com/ivanDonadello/Visual-Relationship-Detection-LTN}. A video showing a demo of the SII system can be seen at \mbox{https://www.youtube.com/watch?v=y2-altg3FFw}.} on the Visual Relationship Dataset (VRD) \cite{lu2016visual} that contains 4000 images for training and 1000 for testing annotated with visual relationships. Bounding boxes are annotated with a label in $\P_1$ containing 100 unary predicates. These labels refer to animals, vehicles, clothes and generic objects. Pairs of bounding boxes are annotated with a label in $\P_2$ containing 70 binary predicates. These labels refer to actions, prepositions, spatial relations, comparatives or preposition phrases. The dataset has 37993 instances of visual relationships and 6672 types of relationships. 1877 instances of relationships occur only in the test set and they are used to evaluate the zero-shot learning scenario.

\paragraph{VRD Tasks} The performance of LTNs are tested on the following VRD standard tasks. The \textbf{phrase detection} is the prediction of a correct triple $\langle subject$, $predicate$, $object \rangle$ and its localization in a single bounding box containing both the subject and the object. The triple is a true positive if the labels are the same of the ground truth triple and if the predicted bounding box has at least 50\% of overlap with a corresponding bounding box in the ground truth. The ground truth bounding box is the union of the ground truth bounding boxes of the subject and of the object. The \textbf{relationship detection} task predicts a correct triple/relationship and the bounding boxes containing the subject and the object of the relationship. The triple is a true positive if both bounding boxes have at least 50\% of overlap with the corresponding ones in the ground truth. The labels for the predicted triple have to match with the corresponding ones in the ground truth. The \textbf{predicate detection} task predicts a set of correct binary predicates between a given set of bounding boxes. Here, the prediction does not depend on the performance of an object detector. The focus is only on the ability of LTNs to predict binary predicates.

\paragraph{Comparison} The performance of LTNs on these tasks have been evaluated with two LTNs grounded theories (or models): $\gtwo$ and $\gtw$. In the first one, $\gtwo = \langle \kwo, \pgsii \rangle$, $\kwo$ contains only positive and negative examples for the predicates in $\P$. This theory gives us the first results on the effectiveness of LTNs on visual relationship detection with respect to the state-of-the-art. In the second grounded theory $\gtw = \langle \kw, \pgsii \rangle$, $\kw$ contains examples and the logical constraints. With $\gtw$ we check the contribution of the logical constraints w.r.t. a standard machine learning approach. We first train the LTNs models on the VRD training set and then we evaluate them on the VRD test set. The evaluation tests the ability of LTNs to generalize to the 1877 relationships never seen in the training phase.

Before comparing the two models with the state-of-the-art, we perform some ablation studies to see what are the key components of our SII system based on LTNs (see Table \ref{tb:abl}). A first key feature are the logical constraints in $\gtw$ w.r.t. $\gtwo$. The second component is the contribution of the new joint features between bounding boxes, Equation \eqref{eq:grCC}. These features represent a novelty of our work w.r.t. the classical features of the LTNs framework. The third important aspect is the adoption of a loss function based on the harmonic mean of the clauses in $\ksii$, see Equation \eqref{eq:ltn_loss_sii}. This differs from the classical LTNs loss function based on the t-norm (e.g., the minimum) of the clauses in $\ksii$. We test $\gtwo$ and $\gtw$ by adding and removing the new features and by adopting the new loss function instead of the classical one of LTNs.

The LTNs models $\gtwo$ and $\gtw$ are then compared with the following methods of the state-of-the-art, see Table \ref{tb:vrd_results_zero}. VRD \cite{lu2016visual} is the seminal work on visual relationship detection and provides the VRD. The method detects visual relationships by combining visual and semantic information. The visual information is the classification score given by two convolutional neural networks. The first network classifies single bounding boxes according to the labels in $\P_1$. The second one classifies the union of two bounding boxes (subject and object) according to the labels in $\P_2$. These scores are combined with a language prior score (based on word embeddings) that models the semantics of the visual relationships. The methods in \cite{tresp2017improvingVRD} also combines visual and semantic information. However, link prediction methods (\emph{RESCAL}, \emph{MultiwayNN}, \emph{CompleEx}, \emph{DistMult}) are used for modelling the visual relationship semantics in place of word embeddings. In LKD \cite{yu17vrd} every visual relationship is predicted with a neural network trained on visual features and on the word embeddings of the subject/object labels. This network is regularized with a term that encodes statistical dependencies (taken from the training set and Wikipedia) between the predicates and subjects/objects. In VRL \cite{Liang2017Deep} the semantic information of the triples is modelled by building a graph of the visual relationships in the training set. A visual relationship is discovered (starting from the proposals coming from an object detector) with a graph traversal algorithm in a reinforcement learning setting. Context-AwareVRD \cite{Zhuang2017context} and WeaklySup \cite{Peyre2017Weakly} encode the features of pairs of bounding boxes similarly to Equation \eqref{eq:grCC}. However, in Context-AwareVRD the learning is performed with a neural network, whereas in WeaklySup the learning is based on a weakly-supervised discriminative clustering I.e., the supervision on a given relationship is not at triples level but on an image level. 

\paragraph{Evaluation Metric}
For each image in the test set, we use $\gtwo$ and $\gtw$ to compute the ranked set of groundings $\{\pgsii(\mathsf{r}(b,b'))\}_{\mathsf{r} \in \P_2}$, with $\langle b, b' \rangle$ bounding boxes computed with an object detector (the R-CNN model in \cite{lu2016visual}\footnote{https://github.com/Prof-Lu-Cewu/Visual-Relationship-Detection}) or taken from the ground truth (for the predicate detection). Then we perform the post processing. As metrics we use the \textbf{recall@100/50} \cite{lu2016visual} as the annotation is not complete and precision would wrongly penalize true positives. We classify every pair $\langle b, b' \rangle$ with \emph{all} the predicates in $\P_2$ as many predicates can occur between two objects (e.g., a person rides and is on a horse at the same time) and it is not always possible to define a preference between predicates. This choice is counterbalanced by predicting the correct relationships within the top 100 and 50 positions.

\paragraph{Implementation Details}
In Equations \eqref{eq:grUniv} and \eqref{eq:ltn_loss_sii} we set $p=-1$ (harmonic mean). The chosen t-norm is the \L{}ukasiewicz one. The number of tensor layers in Equation \eqref{eq:grP} is $k=5$ and $\lambda=10^{-10}$ in Equation \eqref{eq:ltn_loss_sii}. The optimization is performed separately on $\kw$ and $\kwo$ with 2500 training epochs of the RMSProp optimizer in \tensorflow.

\begin{ijcnn}
\subsection{Ablation Studies}
\begin{table*}[!h]
\centering
\caption{Ablation studies for the LTNs models. The combination of the harmonic-mean-based loss function (hmean) and the new features (newfeats) leads to the best results for both $\gtwo$ and $\gtw$.}
\label{tb:abl}
\begin{tabular}{lcccccc}
\hline\hline
Task                      & Phrase Det.                 & Phrase Det.                 & Relationship Det.           & Relationship Det.           & Predicate Det.              & Predicate Det.              \\\hline
Evaluation                & R@100                       & R@50                        & R@100                       & R@50                        & R@100                       & R@50                        \\\hline\hline
$\gtwo$ (min)             & 10.04  $\pm$ 0.46 & 6.15  $\pm$ 0.28  & 9.11  $\pm$ 0.41  & 5.56  $\pm$ 0.24  & 68.06  $\pm$ 0.67 & 48.80  $\pm$ 1.04 \\
$\gtw$ (min)              & 10.21  $\pm$ 0.63 & 6.29  $\pm$ 0.45  & 9.31  $\pm$ 0.58  & 5.70  $\pm$ 0.39  & 67.83  $\pm$ 0.80 & 48.41  $\pm$ 0.84 \\
$\gtwo$ (min, newfeats)   & 12.24  $\pm$ 0.36 & 8.35  $\pm$ 0.51  & 11.20 $\pm$ 0.38 & 7.60  $\pm$ 0.46  & 69.91  $\pm$ 0.78 & 51.40  $\pm$ 0.76 \\
$\gtw$ (min, newfeats)    & 11.92  $\pm$ 0.34 & 8.15  $\pm$ 0.45  & 10.90 $\pm$ 0.34 & 7.47  $\pm$ 0.44  & 69.83  $\pm$ 0.72 & 51.22  $\pm$ 1.01 \\
$\gtwo$ (hmean)           & 13.35  $\pm$ 0.56 & 8.97  $\pm$ 0.50  & 12.22 $\pm$ 0.53 & 8.12  $\pm$ 0.46  & 71.11  $\pm$ 0.75 & 51.36  $\pm$ 0.91 \\
$\gtw$ (hmean)            & 13.63  $\pm$ 0.41 & 9.25  $\pm$ 0.45  & 12.52 $\pm$ 0.46 & 8.49  $\pm$ 0.44  & 72.68  $\pm$ 0.52 & 52.78  $\pm$ 0.36 \\
$\gtwo$ (hmean, newfeats) & 15.91  $\pm$ 0.54 & 11.00 $\pm$ 0.34 & 14.65  $\pm$ 0.54 & 10.01  $\pm$ 0.38 & 74.71  $\pm$ 0.73 & 56.25  $\pm$ 0.75 \\
$\gtw$ (hmean, newfeats)  & 15.74  $\pm$ 0.44 & \textbf{11.40 $\pm$ 0.31} & 14.43  $\pm$ 0.41 & \textbf{10.47  $\pm$ 0.32} & \textbf{77.16  $\pm$ 0.60} & \textbf{57.34  $\pm$ 0.78} \\\hline\hline 
\end{tabular}
\end{table*}
Table \ref{tb:abl} shows the results of the ablation studies. We perform the training 10 times obtaining 10 models for $\gtwo$ and $\gtw$, respectively. For each task and for each grounded theory we report the mean and the standard deviation of the results given by these models. In addition, we perform the t-test with p-value 0.05 to statistically compare the LTNs models in each task. The first LTNs models use the classical LTNs loss function without the new joint features. Their performance are statistically improved of approximately 2 points in all the tasks if we add the new joint features. However, the minimum t-norm (i.e., the G\"odel t-norm) leads the optimization process to a local optimum, as stated in Section \ref{sec:opt_sii}, thus vanishing the effect of the logical constraints. Indeed, $\gtwo$ and $\gtw$ have similar performance by adopting the G\"odel t-norm\footnote{The other t-norms lead to a non-converging optimization process due to the numerical issues mentioned in Section \ref{sec:opt_sii}.}. If we adopt the harmonic-mean-based loss function we can see a statistical improvement of the performance of both $\gtwo$ (hmean) and $\gtw$ (hmean). These results are further statistically improved by adding the new features, see results for $\gtwo$ (hmean, newfeats) and $\gtw$ (hmean, newfeats). This novelty is fundamental as it allows us to prove the contribution of the logical constraints. Indeed, $\gtw$ has statistically better performance (in bold) of $\gtwo$ in the predicate detection. In the other tasks the improvement can be statistically observed for the recall@50, indeed the logical constraints improve the precision of the system. The introduction of the new joint features and the new loss function gives an improvement of the recall of approximately 9 points for the predicate detection task and of approximately 5 points for the other tasks.
\end{ijcnn}

\subsection{Comparison with the State-of-the-Art}
\label{sec:ltn_vrd_zero}
Table \ref{tb:vrd_results_zero} shows the LTNs results compared with the state-of-the-art. The phrase and the relationship detection tasks are the hardest tasks, as they include also the detection of the bounding boxes of the subject/object. Therefore, the errors coming from the object detector propagate also to the visual relationship detection models. Adopting the same object detector used by VRD, \cite{tresp2017improvingVRD} and WeaklySup allows us to compare LTNs results starting from the same level of error coming from the bounding boxes detection. The $\gtwo$ and $\gtw$ models outperform these competitors showing that LTNs deal with the object detection errors in a better way. The predicate detection task, instead, is easier as it is independent from object detection. In this task, it is possible to see all the effectiveness of LTNs due to the good improvement of performance. Regarding the other competitors, a fully comparison is possible only in the predicate detection task as in the other tasks they use a different object detector for bounding boxes detection. However, both LTNs models effectively exploit the structure of the data to infer similarity between relationships and outperform VRL and Context-AwareVRD. Moreover, the LTNs model $\gtw$ trained with data and constraints outperforms the model $\gtwo$ trained with only data. This means that the negative domain and range constraints are effective at excluding some binary predicates for a bounding box with a given subject or object. For example, if the subject is a physical object, then the predicate cannot be $\sleepOn$. This peculiarity of LTNs is fundamental in the comparison with LKD, as $\gtw$ achieves the improvement that outperforms LKD on predicate detection. The performance on the other tasks are comparable even if the data coming from the object detection are different. Indeed, LKD uses a more recent object detector with better performance than the one provided by VRD. LKD exploits statistical dependencies (co-occurrences) between subjects/objects and relationships, whereas LTNs exploit logical knowledge. This can express more information (e.g., positive or negative dependencies between subjects/objects and relationships, properties of the relationships or relationships) and allows a more accurate reasoning on the visual relationships. Moreover, LKD has to predict $\mathcal{O}(|\P_1|^2|\P_2|)$ possible relationships, whereas the searching space of LTNs is $\mathcal{O}(|\P_1| + |\P_2|)$. This implies an important reduction of the parameters of the method and a substantial advantage on scalability.

\begin{table*}[htbp]
\caption{Results on the Visual Relationship Dataset (R@N stands for recall at N). The use of the logical constraints in $\gtw$ leads to the best results and outperforms the state-of-the-art in the predicate detection task.}
\label{tb:vrd_results_zero}
\centering
\begin{tabular}{lcccccc}
\hline\hline
Task                                    	& Phrase Det. 	& Phrase Det. 	& Relationship Det. 	& Relationship Det. 	& Predicate Det. 	& Predicate Det. 	\\ \hline
Evaluation                              	& R@100       	& R@50        	& R@100     	& R@50     	& R@100      	& R@50       	\\\hline\hline
VRD \cite{lu2016visual}                                                                                        & 3.75  & 3.36  & 3.52  & 3.13  & 8.45  & 8.45  \\
RESCAL \cite{tresp2017improvingVRD}                                                                                          & 6.59  & 5.82  & 6.07  & 5.30  & 16.34 & 16.34 \\
MultiwayNN \cite{tresp2017improvingVRD}                                                                                      & 6.93  & 5.73  & 6.24  & 5.22  & 16.60 & 16.60 \\
ComplEx \cite{tresp2017improvingVRD}                                                                                         & 6.50  & 5.73  & 5.82  & 5.05  & 15.74 & 15.74 \\
DistMult \cite{tresp2017improvingVRD} &4.19	&3.34	&3.85	&3.08	&12.40	&12.40\\
$\gtwo$ (hmean, newfeats) & 15.91  $\pm$ 0.54 & 11.00 $\pm$ 0.34 & 14.65  $\pm$ 0.54 & 10.01  $\pm$ 0.38 & 74.71  $\pm$ 0.73 & 56.25  $\pm$ 0.75 \\
$\gtw$ (hmean, newfeats)  & 15.74  $\pm$ 0.44 & 11.40 $\pm$ 0.31 & 14.43  $\pm$ 0.41 & 10.47  $\pm$ 0.32 & $\mathbf{77.16  \pm 0.60}$  & \textbf{57.34  $\pm$ 0.78} \\
LKD \cite{yu17vrd}                                                                                       & \textbf{17.24} & \textbf{12.96} & \textbf{15.89} & \textbf{12.02} & 74.65 & 54.20  \\
VRL \cite{Liang2017Deep} & 10.31 & 9.17  & 8.52  & 7.94  & -     & -     \\
Context-AwareVRD \cite{Zhuang2017context}                  & 11.30  & 10.78 & 10.26 & 9.54  & 16.37 & 16.37 \\
WeaklySup \cite{Peyre2017Weakly}                                  & 7.80   & 6.80   & 7.40   & 6.40   & 21.60  & 21.60  \\
\hline\hline \\
\end{tabular}
\end{table*}

These results prove two statements: (i) the general LTNs ability in the visual relationship detection problem to generalize to never seen relationships; (ii) the LTNs ability to leverage logical constraints in the background knowledge to improve the results in a zero-shot learning scenario. This important achievement states that it possible to use the logical constraints to compensate the lack of information in the datasets due to the effort of the annotation. These exploited logical constraints are general and can be retrieved from on-line linguistic resources.

\section{Conclusions and Future Work}\label{sec:conclusions}
The zero-shot learning in SII is the problem of detecting visual relationships whose instances do not appear in a training set. This is an emerging problem in AI datasets due to the high annotation effort and their consequent incompleteness. Our proposal is based on Logic Tensor Networks that are able to learn the similarity with other seen triples in presence of logical background knowledge. The results on the Visual Relationship Dataset show that the jointly use of data and logical constraints outperforms the state-of-the-art methods. The rationale is that the logical constraints explicitly state the relations between a given relationship and its subject/object. Therefore, logical knowledge can compensate at the incompleteness of annotation in datasets. In addition, some ablation studies prove the effectiveness of the introduced novelties with respect to the standard LTNs framework. As future work, we plan to apply LTNs to the Visual Genome dataset \cite{krishna2016visualgenome}. We also plan to study the performance by using different categories of constraints. This allows us to check if some constraints more effective than others. In addition, we want to check the robustness of a SII system given by the logical constraints. This can be achieved by training with an increasing amount of flipped labels in the training data as performed during a poisoning attack in adversarial learning \cite{biggio12poisoning}.

\bibliographystyle{plain}
\bibliography{bibliovrd}

\end{document}